\documentclass[10pt, a4paper]{article}

\usepackage[final]{lrec2026} %
\usepackage{microtype}
\usepackage{todonotes}
\usepackage{graphicx}
\usepackage{booktabs}
\usepackage{soul}
\usepackage{enumitem}
\usepackage{expex}
\usepackage{multirow}
\usepackage{makecell}
\usepackage[dvipsnames]{xcolor}
\usepackage[T1]{fontenc}
\usepackage{tabularx}
\usepackage{booktabs}

\title{Conditioning LLMs to Generate Code-Switched Text}

\name{Maite Heredia \quad Gorka Labaka \quad Jeremy Barnes \quad Aitor Soroa} 

\address{HiTZ Center - Ixa, University of the Basque Country UPV/EHU    \\
         maite.heredia@ehu.eus\\
        }

\abstract{
Code-switching (CS) is still a critical challenge in Natural Language Processing (NLP), due to the limited availability of large-scale, diverse CS datasets for robust training and evaluation. Despite recent advances, the capabilities and limitations of LLMs in handling CS are still not fully understood. In this work, we investigate the extent to which LLMs can be used in a framework for CS text generation, focusing on the English-Spanish language pair. Our proposed methodology consists of back-translating natural CS sentences into monolingual English, and using the resulting parallel corpus to fine-tune LLMs to turn monolingual sentences into CS. We thoroughly analyze the models' performance through a study on human preferences, a qualitative error analysis, an evaluation with popular reference-based metrics and LLM-based judgment. Results show that fine-tuning can be a key step to ensure that current LLMs consistently generate fluent code-switched text and that our methodology generates high-quality outputs, expanding research opportunities in CS communication. We find that traditional metrics do not correlate with human judgment, and although LLM-based evaluation aligns somewhat more closely, the agreement remains limited. We release our code and generated dataset under a CC-BY-NC-SA license.\textsuperscript{1}
 \\ \newline  }

\begin{document}

\maketitleabstract
\footnotetext[1]{\href{https://github.com/hitz-zentroa/cs-generation}{Code} \includegraphics[width=0.05\linewidth]{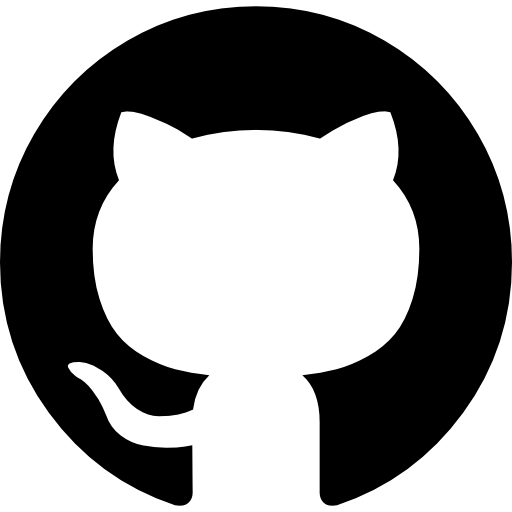} \:\:\:\:\:\href{https://huggingface.co/datasets/HiTZ/EN2CS}{Dataset}\:\includegraphics[width=0.05\linewidth]{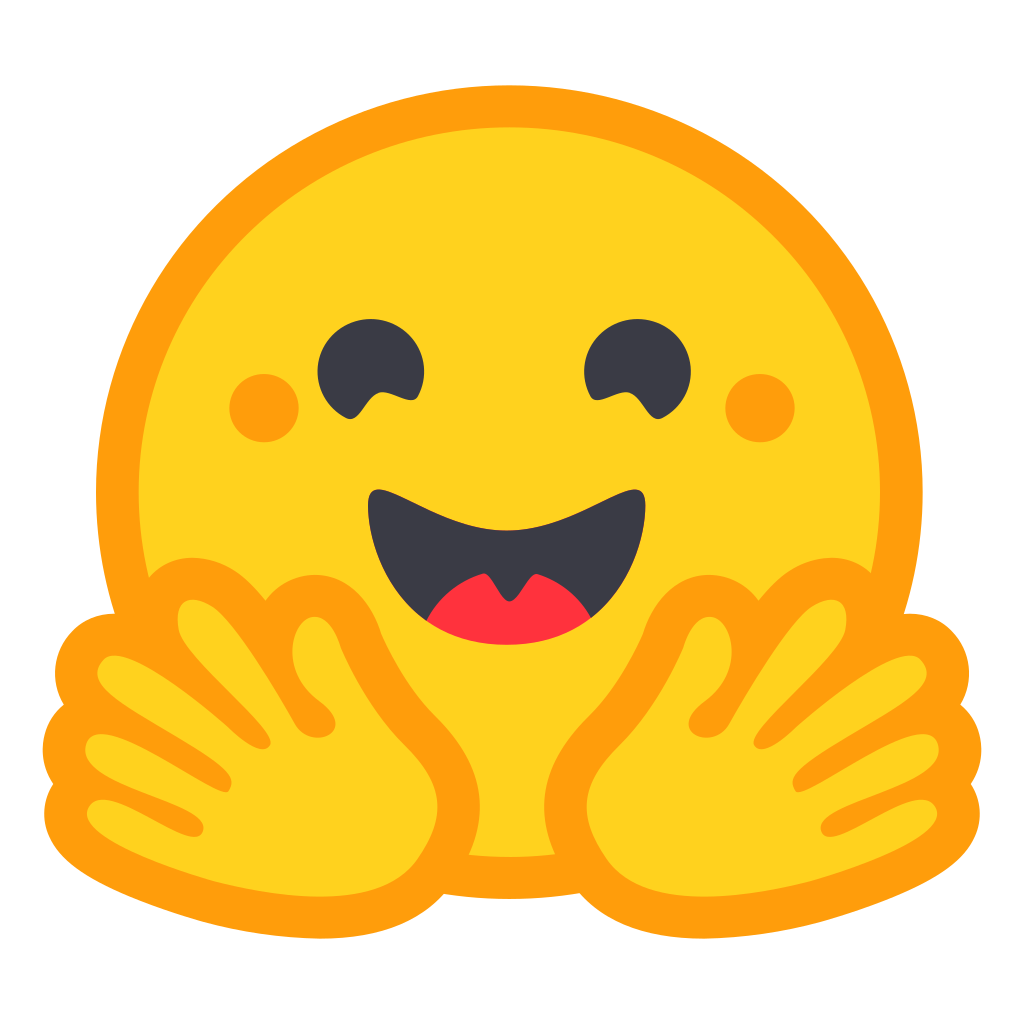}}
\setcounter{footnote}{1}

\section{Introduction}
 \textit{Code-Switching} (CS) consists of mixing two or more languages within a single utterance and is a common phenomenon in multilingual settings \citep{global-bilingualism-tucker}. Although it is mainly present in spoken interactions, it can also be found in written interactions on-line \citep{language-contact-appel-muysken,interlingual-online}, where it appears jointly with other features of informal speech. Example \ref{ex:cs1} shows an utterance where the speaker switches between English and Spanish.

 \ex \label{ex:cs1}
 Why make everybody \textit{sentarse atrás pa' que} everybody has to move \textit{pa' que se salga}.                                                                                                                                                        \\
 Why make everybody \textit{sit at the back so that} everybody has to move \textit{so that she may get out.}\footnote{In all examples of CS featured in this paper, Spanish parts are shown in italics, in both the original instance and its translation.} \\
 \null\hfill\citep{poplack-sometimes}
 \xe

Despite the prevalence of code-switching, most research in Natural Language Processing (NLP) assumes monolingualism as a standard for human communication. However, this implicit decision means that state-of-the-art models are not able to properly interpret or generate CS data. Even advances in \textit{multilingual} language modelling \citep{lin-etal-2022-shot,chowdhery-etal-2023-PaLM} have not led to significant improvements, and performance on CS data is still poor compared to performance on monolingual data \citep{aguilar-etal-2020-lince,winata-etal-2021-multilingual}. This occurs because there is little CS text available in the multilingual pretraining data. Similarly, there are no parallel datasets available to learn to generate CS in a supervised fashion, as one would expect for tasks such as Machine Translation (MT). 
Finally, existing methodology for evaluating automatically generated CS text, which has specific needs different from other text generation tasks, are still not good enough and fail to capture nuances of CS text \mbox{\citep{srivastava-singh-2021-challenges}}. It is therefore crucial to develop methodologies to enable models to generate natural CS text and simultaneously implement robust evaluation frameworks that can assess how well NLP systems handle CS across multiple tasks. We argue that both of these goals require models that can conditionally generate CS from monolingual text.
Consequently, our research focuses on the development of a methodology to fine-tune and evaluate LLMs on the task of CS generation, following three main research questions:
\paragraph{RQ1:} What are the comparative strengths and limitations of fine-tuned versus non-fine-tuned LLMs in generating fluent and natural code-switched text?
\paragraph{RQ2:}How can we leverage LLMs to create high-quality pseudo-parallel data for fine-tuning LLMs in CS text generation? 
\paragraph{RQ3:} Do automatic metrics for Natural Language Generation (NLG) or LLM judges correlate well with human judgment for the task of CS generation?

Based on these research questions, we propose a novel approach to generate CS from monolingual text using LLMs and apply it to the English-Spanish pair. 
We create a new parallel English–CS corpus, \emph{EN-CS}, by leveraging natural CS data and using LLMs to perform back-translation from CS into English, resulting in high-quality pseudo-parallel pairs, suitable for training and evaluating models on CS generation (RQ2). 
We provide a comprehensive comparison of CS generation using LLMs in both zero-shot and fine-tuned settings, and we compare their performance against that of a dedicated MT model (RQ1). 
Finally, we evaluate our methodology both qualitatively, with a study on human preferences and a manual error analysis, and quantitatively, using automatic NLG metrics and LLM as a judge, which allows us to study the correlation between human and automatic evaluation for this task (RQ3). The evaluation is conducted in both in-domain and out-of-domain settings.

\section{Related Work}
\label{sec:related-work}
\paragraph{Perspectives in linguistics.}CS naturally occurs in communities where two or more languages are in contact, making it a subject of interest to fields like sociolinguistics and psycholinguistics. From a social perspective, it can be affected by speakers' attitudes towards the languages and the CS phenomenon itself. In this respect, it is related to notions of prestige and identity \citep{2025-heredia-actitudes}. For example, in bilingual communities where a language is minoritized, CS can be seen as an intrusion of the majority language \citep{dewaele-attitudes}. However, for migrant communities, it may be a way to preserve their mother tongue and as an ``emblem of ethnic identity'' \citep{poplack-sometimes}. Its importance in different social contexts highlights the need to consider CS in NLP research, as it plays a crucial role in linguistic interactions and, consequently, the development of language technologies.

\paragraph{Datasets \& benchmarks for CS.} Most code-switched data stems from social media, while other popular data sources include recordings and transcriptions \citep{winata-etal-2023-decades}. Shared tasks using such CS data have been organized for the tasks of Language Identification \citep{solorio-etal-2014-first-shared-task,molina-etal-2016-overview-second-shared-task} and Sentiment Analysis \citep{patwa-etal-2020-semeval}. Similarly, two benchmarks exist to evaluate model performance on CS text, covering different language pairs and tasks: LINCE \citep{aguilar-etal-2020-lince}, which covers tasks such as Part Of Speech tagging or Sentiment Analysis; and GLUECoS \citep{khanuja-etal-2020-gluecos}, which focuses on NLU tasks for Hindi-English. GLUECoS cannot be currently used without access to the X API.

\paragraph{CS generation.} CS generation has seldom been tackled in previous research. Approaches include linguistically informed techniques to find plausible switching points \citep{pratapa-etal-2018-language-modeling-synthetic,gupta-etal-2020-semi,gregorius-okadome-2022-generating-dependency,hsu-etal-2023-code-switching-synthesis,potter-yuan-2024-llm}, data augmentation techniques \citep{tarunesh-etal-2021-machine-translation} and, more recently, prompting LLMs for CS generation \citep{yong-etal-2023-prompting,terblanche-etal-2024-prompting}. While CS generation is often evaluated by human annotators \citep{tarunesh-etal-2021-machine-translation,gregorius-okadome-2022-generating-dependency}, there remains a need for more robust automatic evaluation methodologies to carefully assess the naturalness and fluency of the generated texts, as recently explored by JudgeLLMs \mbox{\citep{kuwanto2024linguisticstheorymeetsllm}}.

\section{Parallel Data Creation}
\label{sec:method-generate-cs}
In this work we present a novel approach to generate code-switched text from monolingual sentences. As a first step, we create a synthetic parallel corpus from an initial set of English-Spanish CS sentences from the LINCE benchmark~\citeplanguageresource{aguilar-etal-2020-lince} with their English monolingual equivalents, generated by the Command R model~\citeplanguageresource{cohere_for_ai_2024}. We exploit the fact that LLMs struggle to generate CS text given a monolingual sentence (c.f. Section \ref{sec:qual-eval}), but are able to more reliably convert a CS sentence to its corresponding monolingual version, especially when the target language is English. After having created this pseudo-parallel corpus, we use it to fine-tune LLMs on the task of conditional code-switching generation, presented in Section \ref{sec:mono-cs}.
\begin{table}[t]

    \centering
    \resizebox{\columnwidth}{!}{
    \begin{tabular}{lrrrr}
        \toprule
 & \textbf{Train}                                                & \textbf{Dev} & \textbf{Test} &\textbf{Test (ood)}\\
 \cmidrule(lr){2-2}\cmidrule(lr){3-3} \cmidrule(lr){4-4}\cmidrule(lr){5-5} Original & 94,728       & 19,574 & 33,361  &352\\
          Pre-processed                                          & 12,933       & 2,461  & 5,353   &254\\
          \textit{EN-CS}                                         & 10,703       & 791    & 1,040   & 171\\\bottomrule
    \end{tabular}
    }
    \caption{Size of original LINCE (EN-ES) compared to the automatically filtered instances and the final set of parallel instances, dubbed \textit{EN-CS}.}
    \label{tab:parallel-size}
\end{table}
\begin{table*}[t]
    \centering
    \resizebox{\linewidth}{!}{
    \begin{tabular}{lll}
        \toprule
 & \textbf{Original}                          & \textbf{English}                                                                 \\
        \cmidrule(lr){2-2}\cmidrule(lr){3-3}
        \multirow{2}{*}{Silver} 
                                              & you just have to tell me \textit{que como te va}. 
                                              & You just have to tell me \textit{how it's going.}                                \\\cmidrule(lr){2-2}\cmidrule(lr){3-3}
                                              & osea i know we wanna party \textit{pero tampoco no aya asta dallas} 
                                              & like i know we want to party \textit{but not all the way to dallas}              \\
        \midrule
        \multirow{2}{*}{Gold} 
                                              & \textit{hasta venir a plaza se siente} like home. 
                                              & \textit{even coming to the square feels} like home.                              \\\cmidrule(lr){2-2}\cmidrule(lr){3-3}
                                              & \textit{me siento tan pendejo} right now.	
                                              & \textit{i feel so stupid} right now.                                             \\
        \bottomrule
    \end{tabular}
    }
    \caption{Examples of the \textit{EN-CS} parallel corpus. Left: original code-switched instances, right: generated (silver) or post-edited (gold) English instances.}
    \label{tab:parallel-examples}
\end{table*}

\subsection{The LINCE benchmark}
\label{sec:initial-data}
We use LINCE as a starting point, a popular benchmark that has been widely used to evaluate CS systems \citeplanguageresource{aguilar-etal-2020-lince}, which is available in 6 language pairs. 
All sentences in LINCE are tokenized, and each token is annotated with a language tag as well as other categories depending on the task. In our work we focus on the English-Spanish pair and filter all sentences in the data that do not contain CS, similarly discarding all the task-specific annotations. Example \ref{ex:lince} shows a random instance from LINCE.

\ex \label{ex:lince}
\resizebox{0.9\columnwidth}{!}
{%
\begin{tabular}{ccccccc}
         \underline{estaba} & \underline{aquí} & \underline{three} & \underline{feet} & \underline{away} & \underline{.} \\
         spa                & spa              & eng               & eng              & eng              & eng\&spa      \\
\end{tabular} }
\xe

LINCE comprises around $95,000$ train, $20,000$ development, and $33,000$ test instances for the English-Spanish pair. We deduplicate the instances among splits, and filter and pre-process the instances to ensure that they are suitable for our task by removing links, replacing usernames with the placeholder \textit{<user>}, and detokenizing all instances with the script provided as part of the Moses toolkit \citep{koehn-etal-2007-moses}. 
After this preprocessing, we obtain a more natural version of the LINCE data. A preliminary analysis reveals that many sentences in LINCE are monolingual or contain a single word in one language, which often corresponds to a borrowing, as shown in Example \ref{ex:tequila}. We adopt a simple heuristic to approximate the distinction between actual CS and cases that involve only a borrowing or are incorrectly labeled. Specifically, we retain only sentences that contain at least two words in each language, which substantially reduces the likelihood of including cases where an isolated borrowing is tagged as CS.

    \ex  \label{ex:tequila}
        I need a shot of \underline{tequila} or a glass of scotch to keep me warm right now.
    \xe

After these pre-processing and filtering steps, we end up with $12,933$ train, $2,461$ development and $5,353$ test instances. The comparison between the original size of LINCE and the final number of sentences selected for our experiments after pre-processing is shown in Table \ref{tab:parallel-size}.

\subsection{\textit{EN-CS}}
\label{sec:parallel-data-creation}
The next step in our method requires creating a pseudo-parallel English-CS dataset by translating the natural code-switched instances into monolingual text.  As there are no available MT systems to convert from English-Spanish CS text to English monolingual text, we instead make use of prompt engineering, using the Command R model \citeplanguageresource{cohere_for_ai_2024}, one of the strongest publicly available models at the time.

We perform an initial set of experiments to determine the optimal prompt to generate monolingual English versions of the code-switched data. Ideally, we aim for a prompt that generates translations that maintain the meaning of the original sentences, are fluent and natural, whose grammar is correct, and that do not contain any Spanish words or phrases. After extensive testing (see Appendix \ref{sec:app-prompt-tuning}), we use the following prompt in a 5-shot setting: \textit{Now convert this code-switched phrase to English. Leave the parts in English as they are, focus on translating the parts in Spanish.} %
Finally, we filter output instances that contain profanity that was not present in the source texts or irrelevant information, such as \textit{Of course, here's your translation:}, because preliminary experiments show that these instances were problematic for the generation task and conditioned the outputs too much.

In order to create a valid gold standard test set, we perform a manual post-edition of the the monolingual test translations for $1,040$ instances of the LINCE test set. The post-edition was carried out by three proficient speakers of English and Spanish, who were provided with specific guidelines, as shown in Appendix \ref{sec:appendix-post-edition-guidelines}.

Table \ref{tab:parallel-size} shows the final size of the parallel corpus, which we dub \textit{EN-CS}, after post-processing and post-edition, and Table \ref{tab:parallel-examples} shows examples of silver and gold instances. The final version of our dataset therefore contains $10,703$ train and $791$ development instances with automatically translated English sentences matched to their original CS sentences, and $1,040$ gold instances with post-edited English translations.

\begin{table*}[t]
\resizebox{\textwidth}{!}{
    \renewcommand{\arraystretch}{1.2} %
    \centering
    \begin{tabular}{ll}  
        \toprule
        \textbf{Model}                & \textbf{Generated Output}                                                                                                                                                                                     \\ 
        \cmidrule(lr){1-1}\cmidrule(lr){2-2}
        \makecell[l]{Original (Gold)} & damm \textit{todos se casaron }and we still single lol forever alone                                                                                                                                          \\  
        English (Source)              & damn everyone got married and we're still single lol forever alone                                                                                                                                            \\  \cmidrule(lr){1-1}\cmidrule(lr){2-2}
        Llama3                        & damn \textit{todos se fueron a casarse y nosotras estamos solitarias} lol forever alone                                                                                                                       \\  
        Llama3 Instruct               & damm every1 got married and we're still single lol \textit{alonso solit@o} foreverrrr lolololo                                                                                                                \\  
        Llama3.3-70B$_\mathrm{fs}$    & damn \textit{todo el mundo se casó y nosotros seguimos solteros} lol forever alone                                                                                                                            \\
        GPT-4o$_\mathrm{fs}$          & damn \textit{todos se casaron y nosotros seguimos solteros} lol forever alone                                                                                                                                 \\
        NLLB                          & damn everyone got marry and its still single lol forever alone                                                                                                                                                \\
        \bottomrule
    \end{tabular}
    }
    \caption{Example from the test set and the generated outputs of the different models.}
    \label{tab:generated-text}
\end{table*}
\subsubsection{Quality assessment}

We evaluate the quality of the automatic translations (train/dev) by measuring two dimensions: the overall \textit{fluency} of the sentences and \textit{adequacy} of the translations in respect to the source texts. Two fluent English-speaking annotators evaluate the same $100$ random instances using a 5-point Likert scale \mbox{\cite{callison-burch-etal-2007-meta-evaluation-mt}} and obtain $4.6$ (fluency) and $4.5$ (adequacy) points on average, which show the quality of the generated translations. 
A quadratic Cohen's $\kappa$ of $0.57$ indicates moderate agreement, likely due to lower Likert scores (1, 2, and 3) being rarely selected by the annotators, which is a known problem for $\kappa$ \cite{Xu-2014-kappa,barnes2025summ}. In fact, the raw agreement between annotators is substantially higher: $0.71$ for fluency and $0.65$ for adequacy. See Appendix {\ref{sec:IAA}} for further details. 

To estimate the quality of the post-edition process, we compare the post-editions of two annotators on $100$ additional random instances. The results show that $75\%$ of the sentences remain unchanged, as they are already adequate. There is a $87.87\%$ similarity between the post-editions of the two annotators, as measured by Levenshtein distance, demonstrating a high degree of consistency and quality in the post-edition process.

\section{Experimental settings}

\paragraph{Generation settings}
\label{sec:mono-cs}

With \textit{EN-CS} as our starting point, we frame CS generation as a MT task, with English as the source and CS as the target language, where parts of the source sentence have to be translated to Spanish. 
In our experiments, we fine-tune two small-sized generative models from the Llama family, namely, \href{https://huggingface.co/Undi95/Meta-Llama-3-8B-hf}{Llama3 8B} and \href{https://huggingface.co/Undi95/Meta-Llama-3-8B-Instruct-hf}{Llama3 Instruct 8B} \citelanguageresource{dubey2024llama}.

To fine-tune the models, we use the causal language modelling objective, but with appropriate input formats for the base and instruct models. For the base model, we use templates \citep{zhu-etal-2024-multilingual} in the form of ``\texttt{<X>=<Y>}'', where \texttt{<X>} and \texttt{<Y>} are placeholders for the input English sentence and generated CS, respectively. At inference, the second code-switched part is left empty for the model to fill. For the instruction-tuned model, we provide a system prompt with the instruction, a query by the user in English, and an answer from the assistant with the code-switched target. At inference time, the answer is left blank (See Appendix~\ref{sec:app-ft-prompting} for example prompts).

Models are trained using Quantized Low-Rank Adaptation (QLoRA) \citep{dettmers2023qloraefficientfinetuningquantized} with standard parameters: the model is loaded in 4 bit with NF4 quantization data type and bf16 computational data type. The LoRA rank and scaling factor are set to 16 and the dropout to $0.05$. We apply the LoRA update matrices to the attention blocks and do not train bias parameters. Regarding the hyperparameters, we only tune the learning rate ($1e\textsuperscript{-4}$, $5e\textsuperscript{-4}$, $1e\textsuperscript{-3}$ and $5e\textsuperscript{-3}$) and training epoch $\in [1\ldots10]$, choosing the parameters that give the lowest cross-entropy loss on the development set for each model. We use the transformers package \citep{wolf-etal-2020-transformers} for all training experiments.

Early experiments indicated that fine-tuned models usually produce the desired output up to a punctuation mark and then either begin to translate the sentence again or hallucinate more content. We therefore truncate the output up to a punctuation mark where the length is closest to that of the original sentence \citep{bawden-yvon-2023-investigating-translation-bloom}. Although simple, this has proven to be the method that yields the best results. We additionally experimented with different generation parameters ( \textit{length\_penalty} and \textit{exponential\_length\_decay}), as well as trying to control the length of the generation with length codes, but find that the truncation heuristic performs the best. Accordingly, all further experiments will use the truncated output. %

\paragraph{In-domain and out-of-domain evaluation} 
To evaluate the performance of models, we test them on the test set of \textit{EN-CS}, and to evaluate their capabilities outside of the domains covered by LINCE, we also propose an out-of-domain evaluation. For this purpose, we gather a series of small creative non-fiction texts originally written using English-Spanish CS \citeplanguageresource{mybrother,docehoras,gringos,lamanda,simon,lavaca}. These are quite different from the other split of the test set in domain and register, as well as more superficial features such as the length of the sentences, that tend to be much longer than those in LINCE. To process this collection of texts in a similar manner to the instances of LINCE, we first divide them into sentences and then obtain the language ID for each token using the models from the CodeSwitch repository.\footnote{\url{https://github.com/sagorbrur/codeswitch}} With each sentence tokenized and tagged with its corresponding language, we can then process and filter the instances just like those sourced from LINCE (c.f. Section \ref{sec:initial-data}). 

\paragraph{Baselines} We include few-shot experiments by directly prompting GPT-4o and \mbox{\href{https://huggingface.co/meta-llama/Llama-3.3-70B-Instruct}{Llama3.3 70B Instruct}} to generate CS text using a 5-shot approach (See Appendix~\ref{sec:app-ft-prompting} for example prompts). We refer to these systems in the experiments as GPT-4o$_\mathrm{fs}$ and Llama3.3-70B$_\mathrm{fs}$, respectively. We also include a strong dedicated MT baseline, developed by fine-tuning the NLLB \mbox{\cite{team2022NoLL}} model (\texttt{nllb-200-distilled-600M}) using \emph{EN-CS}. The model was trained with standard settings.\footnote{A batch size of $32$, learning rate of $1 \times 10^{-4}$, using constant learning rate schedule with $1{,}000$ warmup steps, a gradient clipping threshold of $1.0$, and a weight decay of $1 \times 10^{-3}$. Training was conducted for $50{,}000 steps$. For evaluation, we selected the checkpoint that achieved the highest BLEU score on the development set.}

 Table \ref{tab:generated-text} shows an example of the outputs of the different models, compared to the original code-switched sentence, and the English monolingual sentence that they received as input. 

\section{Qualitative evaluation}
\label{sec:qual-eval}
As a first step to assess the quality of the outputs produced by the different models, we perform a manual qualitative analysis of the results in two parts: a pairwise tournament-based human evaluation, and an in-depth analysis of the most common errors made by the models and their distribution.

\subsection{Preference based evaluation}
\label{sec:pref-based-eval}

We perform a tournament-based evaluation that allows us to determine the ranking of models in terms of human preference. A total of $1260$ instances are matched against each other, corresponding to the outputs of the five models for $210$ English source sentences, as well as the gold standard reference. The evaluation is conducted pairwise, requiring annotators to choose the best out of two sentences or declare a tie. When choosing the best sentence, annotators do not know the original English sentence, nor which model produced what output. This process results in $210\cdot{6\choose 2} = 3150$ comparisons, and was carried out by $14$ annotators, with each annotator performing at least $100$ random comparisons.

Annotators are provided with a series of criteria to choose between the instances, based on the error analysis described in the next section. They must take into account three main criteria, which must be applied in the following order: a) the presence and naturalness of the CS; b) the content and fluency of the sentences; and c) the orthographical errors of the instances (correct punctuation, presence of typos, etc.). Annotators are furthermore asked to avoid declaring ties, unless completely necessary (e.g., in a case where both sentences are completely monolingual and therefore equally incorrect) to compel them to develop a preference. The complete annotation guidelines are available in Appendix \ref{sec:annotation-guidelines}. Inter-annotator agreement on a subset of $200$ sentence pairs shows substantial agreement for the in-domain set ($\kappa=0.61$) and moderate agreement for the out-of-domain set ($\kappa=0.51$). %

\begin{table}[t]
  \small
  \centering
  \resizebox{\columnwidth}{!}{
  \begin{tabular}{lcccc|cc}
    \toprule
      & \multicolumn{2}{c}{\textbf{In domain}} & \multicolumn{2}{c}{\textbf{Out of domain}}& \multicolumn{2}{c}{\textbf{Total}} \\
    
    \textbf{Model} 
      & \textbf{Score} & \textbf{Rank} 
      & \textbf{Score} & \textbf{Rank} 
      & \textbf{Score} & \textbf{Rank} 
      \\
    \cmidrule(lr){1-1} \cmidrule(lr){2-3} \cmidrule(lr){4-5} \cmidrule(lr){6-7}
    Gold Standard              
      & 369.5& \includegraphics[width=0.05\columnwidth]{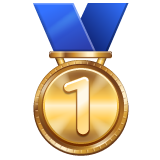} 
      & 434.5& \includegraphics[width=0.05\columnwidth]{figures/first-place.png} 
      & 804.0& \includegraphics[width=0.05\columnwidth]{figures/first-place.png} \\
    Llama3                     
      & 291.5& \includegraphics[width=0.05\columnwidth]{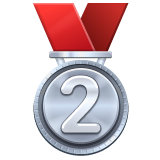} 
      & 282.0& \includegraphics[width=0.05\columnwidth]{figures/second-place.png} 
      & 573.5&       \includegraphics[width=0.05\columnwidth]{figures/second-place.png} \\
    Llama3 Instruct            
      & 270.5& \includegraphics[width=0.05\columnwidth]{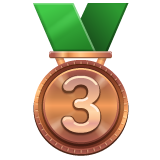} 
      & 210.0& 4& 480.5&       4\\
    NLLB                       
      & 259.5& 4 & 247.5& \includegraphics[width=0.05\columnwidth]{figures/third-place.png}& 507&       \includegraphics[width=0.05\columnwidth]{figures/third-place.png} \\ 
    GPT-4o$_\mathrm{fs}$       
      & 251.5& 5& 162.0& 6 
      & 413.5&       5\\
    Llama3.3-70B$_\mathrm{fs}$ 
      & 207.5& 6& 164.0& 5& 371.5& 6\\
    \bottomrule
  \end{tabular}
  }
  \caption{Ranking of models according to human preference. } 
  \label{tab:human-ranking}
\end{table}

We calculate a global score for each model, as follows: every time a model is voted, it gets $1$ point, and the loser gets $0$ points; in case of ties, both models get $0.5$ points each. Table \ref{tab:human-ranking} shows the global scores, as well as the ranking of human preferences according to said score (second and third columns). We find that the gold standard reference obtains the highest score, as expected, with an specially stark difference in the out-of-domain set, highlighting the difficulty of this domain for all models. Fine-tuned Llama3 ranks the highest among the automatic methods in both in-domain and out-of-domain settings, which shows its ability to generalize to more challenging domains. The instruction-tuned model Llama3 Instruct obtains worse scores compared to its base model counterpart, which is likely due to instruction tuning reducing certain model capabilities \cite{Li2024PredictingVA,west2025base}. It ranks higher than the NLLB model in-domain, but lower in the out-of-domain evaluation and the overall scores. Still, the difference between Llama3 and NLLB shows the potential to rival dedicated models in generation tasks.
According to these preferences, fine-tuning LLMs for CS generation can be critical to ensure better results, since the larger models with few-shot prompting rank, both GPT-4o$_\mathrm{fs}$ and Llama3.3-70B$_\mathrm{fs}$, are outranked by all models in both evaluation settings.

\begin{figure*}[t]
    \centering
        \includegraphics[width=1\textwidth]{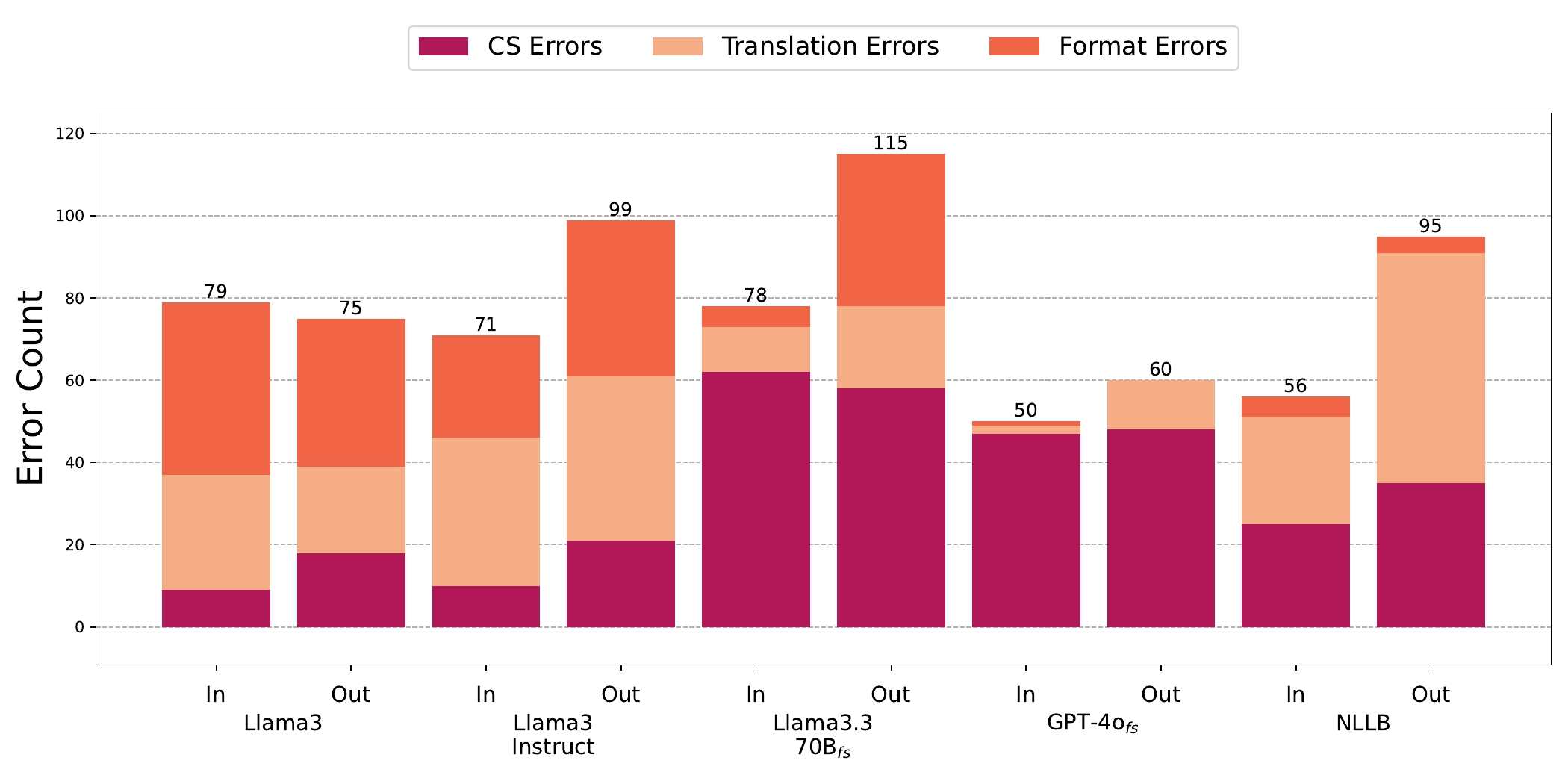}
    \caption{Error distribution by model, obtained by counting the number of instances that present errors of each type. Analyzed both for the in-domain (In) and out-of-domain (Out) sets.
    }
    \label{fig:error-analysis}
\end{figure*}
\subsection{Error analysis}
\label{sec:error-analysis-1}

In order to further explore differences between model performance, we analyze the most common errors made by the CS generation models, both quantitatively and qualitatively. We extend the MT error typology presented in \citet{Popović2018errorclassification} to CS generation error analysis. To do so, we randomly select a set of $100$ outputs from all models and conduct a detailed examination of the types of errors present in them. This thorough analysis allows us to identify recurring patterns and propose a refined error typology specifically for automatic CS generation. This initial error analysis yields $18$ total error categories, which we simplify and group into three main error types: a) CS errors, b) Translation Errors, and c) Format errors. The full error typology, along with detailed descriptions for each error type, is provided in Appendix \ref{sec:appendix-error-typology}.

\begin{itemize}
    \item[] \textbf{CS Errors}: Errors of sentences that are either completely monolingual or switch between languages in an unnatural manner, e.g., by repeating the same word in English and Spanish. In Example \ref{ex:error1}, Llama3.3-70B$_\mathrm{fs}$ preserves the original meaning, but the sentence is fully monolingual.

    \vspace*{-.2cm}
    \ex \label{ex:error1}
    \begin{tabular}{p{0.07\textwidth} p{0.7\linewidth}}
         \textbf{Source} & yea... the best i can do is be here for him if he needs me\\
         \textbf{Output} & \textit{sí... lo mejor que puedo hacer es estar aquí para él si me necesita}\\
    \end{tabular} 
    \xe
    \vspace*{-1cm}

    \item[] \textbf{Translation errors}: Critical errors that either change the original meaning of the sentence or introduce mistakes in fluency or grammar, for example, using the wrong tense or word order. Example \ref{ex:error2} shows an instance where Llama3 Instruct outputs a seemingly natural code-switched sentence, but the phrase ``they got hurt'' is not adequately translated and the meaning of the sentence is not preserved.

    \vspace*{-.2cm}
    \ex\label{ex:error2}
    \begin{tabular}{p{0.07\textwidth} p{0.7\linewidth}}
         \textbf{Source} & I wasn't happy because they got hurt.                  \\
         \textbf{Output} &    i wasn't happy because \textit{me dolieron} \\
    \end{tabular} 
    \xe
    \vspace*{-1cm}
    
    \item[] \textbf{Format errors}: Errors in form that do not make the sentences unintelligible nor change their meaning, such as repetitions of a word or phrase or incorrect punctuation. Example \ref{ex:error3}, by the model Llama3, accurately preserves the original meaning and introduces CS, but removes the username and adds a smiley face.
    
    \vspace*{-.2cm}
    \ex\label{ex:error3}
    \begin{tabular}{p{0.07\textwidth} p{0.7\textwidth}}  
         \textbf{Source} & <user> old mexican remedies               \\
         \textbf{Output} & old school \textit{remedios mexicanos} :) \\
    \end{tabular} 
    \xe
    \vspace*{-1cm}
   
\end{itemize}

We classify $1,000$ additional instances ($200$ instances per model, obtained from the same source sentences) into these kind of errors, and show the results in Figure \ref{fig:error-analysis}. 

For the in-domain test set, GPT-4o$_\mathrm{fs}$ makes the fewest errors overall ($50$), closely followed by NLLB ($56$). However,  $90\%$ of GPT's errors ($45$) and $45\%$ on NLLB's ($25$) are CS related, indicating that while these systems preserve the meaning of sentences and generate few formatting errors, they often produce entirely monolingual outputs, which is a critical error. 
In comparison, CS-related mistakes are the least common in fine-tuned LLMs, accounting for less than $15\%$ of the overall error count. This analysis shows that fine-tuned LLMs have effectively learned to switch between languages naturally, though they may still be prone to other less critical types of errors. The Llama3 base model struggles with maintaining the format of the sentences, which makes up $50.68\%$ of its errors, whereas the instruction-tuned model Llama3 Instruct presents more meaning-related issues, $53,45\%$. This suggests that the linguistic knowledge of the model has degraded when tuned on instructions, a phenomenon that has been observed on other related areas~\cite{fu-etal-2024-disperse}. 
 \begin{table*}[t]
\centering
\resizebox{\textwidth}{!}
{%
\begin{tabular}{lcccccc}
\hline
    & \multicolumn{2}{c}{\textbf{BLEU}}& \multicolumn{2}{c}{\textbf{BERTScore}}& \multicolumn{2}{c}{\textbf{chrF}}\\
 \textbf{Model}&  In domain&Out of domain&  In domain&Out of domain& In domain&Out of domain\\ 
  \cmidrule(lr){1-1}\cmidrule(lr){2-3} \cmidrule(lr){4-5}\cmidrule(lr){6-7}Llama3 8B & \underline{34.49}  &14.93& 81.64&75.44& \underline{53.17}  &38.26\\
  Llama3 8B Instruct                                                                 & 33.42              &14.59& 81.77               &74.99& 52.01              &\underline{38.62}\\ \hline
 Llama3.3-70B$_\mathrm{fs}$                                                          & 22.41              &14.10& 79.77               &78.33& 44.57              &37.37\\
 GPT-4o$_\mathrm{fs}$                                                                & 32.25              &\underline{15.65}& \underline{83.09}   &\underline{78.96}& 50.48              &\underline{38.62}\\
 NLLB                                                                                & \textbf{35.56}     &14.45& \textbf{84.11}      &76.53& \textbf{54.74}     &38.38\\ \hline
 Identity                                                                            & 33.34              &\textbf{41.54}& 82.31               &\textbf{83.31}& 45.51              &\textbf{58.03}\\
 \hline
\end{tabular}%
}
\caption{Results of reference-based metrics the \textit{EN-CS} test set. Best results in bold, second best results underlined.}
\label{tab:test-results}
\end{table*}
When evaluated on a different domain, the Llama3 Instruct and NLLB models produce notably more errors, suggesting that they are overfitting on the train set or have limited ability to generalize to more challenging domains. In particular, NLLB duplicates the number of translation errors. It is also noteworthy that Llama3.3 70B$_\mathrm{fs}$ shows a similar increase in errors, despite not being fine-tuned---a behavior that seems to arise due to the domain and longer sentences. This is not the case for Llama3 or GPT-4o$_\mathrm{fs}$, which output a similar number of errors for both domains. It is especially interesting that the Llama3 base model generalizes so well, despite extensive fine-tuning. Still, the small size of the out-of-domain test set calls for caution when interpreting these differences.

\section{Automatic Evaluation}
\label{sec:automatic-evaluation}

Previous research highlights the challenges of automatically evaluating code-switching (CS) generation, with many existing metrics showing low correlation with human judgments \citep{srivastava-singh-2021-challenges,kuwanto2024linguisticstheorymeetsllm}. On the other hand, recent studies show that using LLMs as evaluators or judges can offer a promising alternative to evaluate generation tasks, as they show a higher alignment with human ratings \mbox{\citep{chiang-lee-2023-large-alternative-evaluations,wang-etal-2023-chatgpt-evaluator}}. In this section, we present the results of an automatic evaluation with reference-based NLG metrics (BLEU, BERTScore, chrF) and GPT-4o as a judge, as well as their correlation with human preferences.

\subsection{Reference-based metrics}
\label{sec:automatic-metrics}
We report the results of BLEU \citep{papineni-etal-2002-bleu}, BERTScore \citep{zhang2020bertscoreevaluatingtextgeneration}, and chrF \citep{popovic-2015-chrf}. \footnote{The metrics are implemented using the \href{https://huggingface.co/docs/evaluate/index}{evaluate} library.} All three are task-agnostic quality metrics that give results between 0-1, based on character-level F-score, n-gram precision and semantic similarity using contextual embeddings\footnote{BERTscore has been calculated using the embeddings from the model \href{https://huggingface.co/google-bert/bert-base-multilingual-cased}{Bert Base Multilingual Cased}.} respectively. We compute the metrics for all systems, and include an Identity system that simply returns the provided input as the output.

The results of the evaluation can be seen in Table \ref{tab:test-results}. 
In the in-domain setting, the best model is NLLB, with the highest scores for the three metrics. It is closely followed by GPT-4o$_\mathrm{fs}$, with the second highest BERTScore, and fine-tuned Llama3, with the second highest BLEU and chrF. Llama3 Instruct follows closely. The Identity system scores nearly as well as the top-performing models for the in-domain set and obtains the best scores out-of-domain. The strong results from the Identity baseline and few-shot models, which often produce monolingual outputs, suggest that reference-based metrics assign high scores to models that match only the English part of the reference. This reflects the nature of the task and dataset, and highlights the limitations and artifacts of using reference-based metrics to evaluate code-switched generation. 
Comparing these results with the out-of-domain evaluation, we observe that all models achieve considerably lower scores. Although GPT-4o$_\mathrm{fs}$ obtains the highest scores, the margin is relatively small. This drop in performance is likely due to specific characteristics of the out-of-domain dataset that negatively affect all models’ predictions, such as longer sentence lengths.

\subsection{GPT as a judge}
\label{sec:gpt-as-judge}

As a complementary automatic assessment of the outputs of our models, we have implemented a zero-shot pair-wise evaluation using GPT-4o as a judge, mimicking the settings of the human evaluation. Details about the implementation are included in Appendix~{\ref{sec:appendix-gpt-judge}.}
Results are shown in Table~{\ref{tab:gpt-judge}}. 
\begin{table}
    \centering
    \begin{tabular}{lcc}\toprule
          \multicolumn{3}{c}{\textbf{GPT}}\\
          \textbf{Model}&\textbf{Score} &\textbf{Rank}\\\cmidrule(lr){1-1}\cmidrule(lr){2-3}
          Gold Standard&744.5&  \includegraphics[width=0.05\columnwidth]{figures/first-place.png} \\
          GPT-4o$_\mathrm{fs}$&603.0&        \includegraphics[width=0.05\columnwidth]{figures/second-place.png}\\
          Llama3.3-70B$_\mathrm{fs}$ &582.0&  \includegraphics[width=0.05\columnwidth]{figures/third-place.png}\\
          Llama3 8B&440.5&        4\\
          Llama3 8B Instruct&410.0&  5\\
          NLLB&370.0&        6\\ \bottomrule

    \end{tabular}
    \caption{Ranking of models using GPT as a judge.}
    \label{tab:gpt-judge}
\end{table}
GPT shows a strong preference for few-shot models, whereas these models are ranked last and second-to-last by humans. 
Based on the error analysis (c.f Section~{\ref{sec:error-analysis-1}}), few-shot models tend to make many CS-related mistakes, as they often output purely monolingual sentences, although they are the most fluent overall.
This suggests that this disagreement may be caused by the humans adhering to the guidelines and taking the presence of CS as the main criterion, whereas GPT is making decisions based on the style and fluency of the answers. Regarding fine-tuned LLMs, they share the same relative ranking in both evaluations, with Llama3 being the preferred model by humans and GPT. Finally, NLLB is ranked last by GPT, while it is the second best model overall as ranked by humans. Further research is needed to explain this behaviour, but there may be some stylistic features of the NLLB model's outputs that are affecting GPT's preferences.

\subsection{Correlation With Human Evaluation}
\label{sec:corr-with-autom-1}

The reference-based metrics used in Section~{\ref{sec:automatic-metrics}} are known to have weak correlations with human judgment in NLG tasks \mbox{\citep{survey-nlg-metrics}}, whereas JudgeLLM-based evaluations seem promising~\mbox{\citep{chiang-lee-2023-large-alternative-evaluations,wang-etal-2023-chatgpt-evaluator}}.
In this section, we compare reference-based metrics and GPT scores with the preference-based scores obtained in Section \ref{sec:pref-based-eval}, allowing us to examine how closely these automatic metrics align with human preferences.
\begin{figure}[t]
    \centering
    \includegraphics[width=\linewidth]{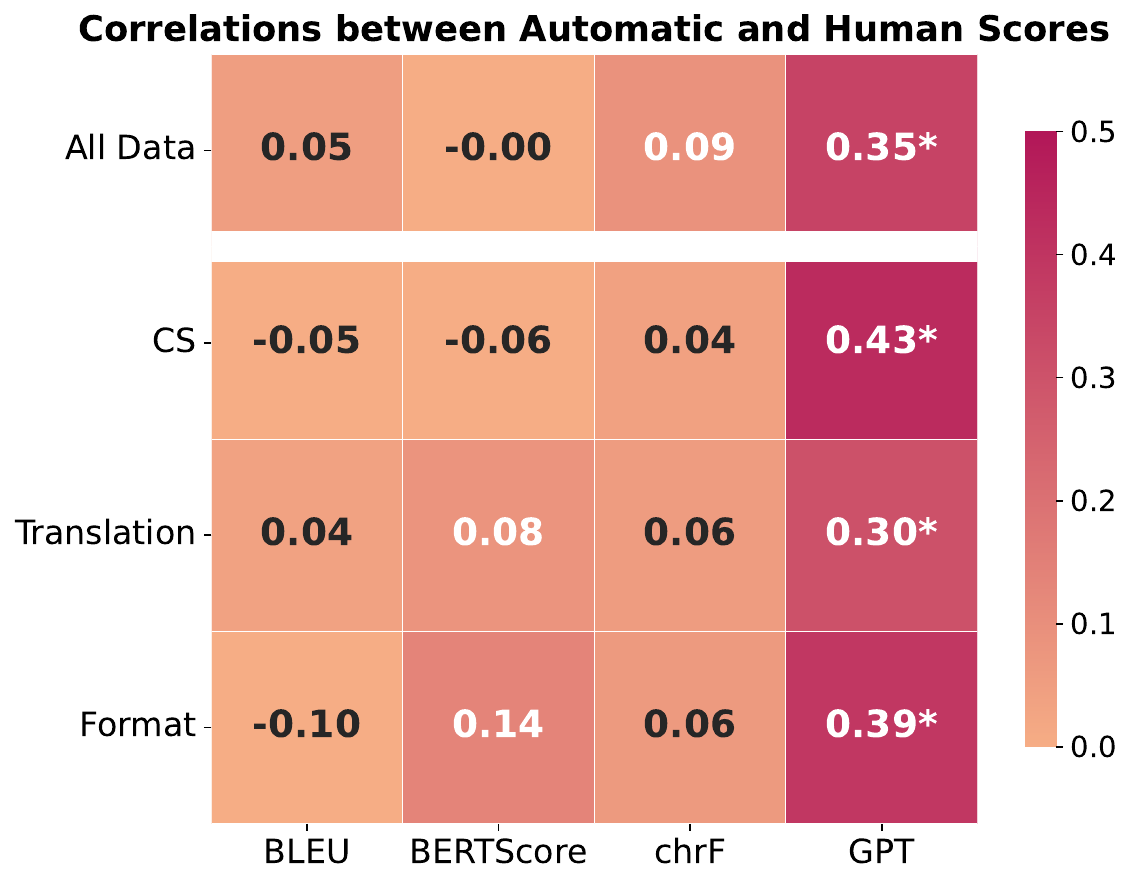}
    \caption{Heatmap of the correlations between human scores and reference-based metrics and scores given by GPT, calculated using the Pearson Correlation Coefficient. The correlations are calculated for all instances, as well as for different subsets of instances, according to the type of errors they exhibit. * indicates statistical significance \((p\le0.05)\).
    }
    \label{fig:heatmap}
\end{figure}

We calculate Pearson's ($\rho$) correlation coefficient at instance-level, using the $1,000$ instances employed for the error classification and human evaluation (the output of $5$ models for $200$ source sentences).\footnote{We do not consider the reference CS sentences when calculating the correlations.} Each data point corresponds to the CS output of one particular model for an English source sentence, and we compute the correlation using two values: the score obtained by the model for this instance in the human preference-based evaluation of Section \ref{sec:pref-based-eval}, and the score it attains if we apply the same strategy using the values of the reference-based metrics to determine the winner, or, in case of JudgeLLM, the scores given by GPT. This allows us to directly compare the two types of evaluations on an instance-by-instance basis.

The correlation coefficients are shown in Figure \ref{fig:heatmap}. The top part of the figure shows the correlation using all the instances, whereas the bottom part only considers those instances that showed some type of error, according to the error analysis described in Section \ref{sec:error-analysis-1}. 
These results provide a more detailed view of how the metrics behave across different subsets of instances.
If we consider all the instances, the maximum $\rho$ correlation value with reference-based metrics is $0.09$, which indicates a low alignment with human scores. GPT-4o$_\mathrm{fs}$ shows a $\rho$ of $0.35$, which is stronger than reference-based metrics, but still too weak to be regarded as a reliable measure for assessing CS generation.

If we instead consider instances with errors from the human evaluation, there is again a higher correlation between human scores and GPT's judgments, with a margin of at least $0.25$ points. Instances with CS errors show the lowest overall correlation with reference-based metrics. This likely derives from the fact that human evaluators never prefer an instance without CS as instructed in the guidelines, but reference-based metrics are not sensitive to these nuances, and may assign high scores to instances regardless of whether they contain CS or not. GPT, however, obtains the highest correlation in these kinds of instances.

All in all, these results confirm that several of the most commonly used reference-based metrics for NLG have a weak correlation with human judgments when evaluating CS generation. This underscores the need to research more specialized evaluation methods designed specifically to capture the nuances of this task that correlate with human judgment.

\section{Conclusion}
In this work, we have presented a methodology to leverage LLMs in the generation of code-switched text from monolingual instances, specifically for the English-Spanish language pair.

Our framework consists of back-translating natural code-switched instances (EN-ES) into monolingual English sentences, and using the resulting parallel corpus, dubbed \textit{EN-CS}, to fine-tune autoregressive models to translate monolingual sentences into CS. This has the advantage of ensuring that the target sentences contain completely natural CS, which has the potential to improve the naturalness of CS generation.

We experiment with fine-tuning base and instruction-tuned LLMs on our dataset using LoRA. For baselines, we include few-shot LLMs (Llama3.3-70B and GPT-4o) and a pretrained NLLB translation system that we also finetune using our dataset. The results indicate that fine-tuned LLMs show higher ranking in a human preference-based evaluation and fewer critical errors than the other baselines, performing better even than proprietary models such as GPT-4o.

We also perform a meta-evaluation of reference-based NLG metrics commonly used for CS evaluation, as well as an LLM judge (GPT-4o). Our analyses show low correlation between human and reference-based evaluations, while the LLM judge achieves moderate correlations. However, particularly in cases with CS errors, no metric is adequate for assessing CS generation. We therefore advocate for more research in specialized evaluation methods. %

\section*{Limitations}
Our research focuses on testing the capabilities of LLMs for CS generation, a field of interest in the research of many applications, yet still in need of more research. While our findings highlight promising potential, we also identify key areas for refinement and improvement, as well as promising lines for future research in this domain.

We want to acknowledge the fact that our approach is dependent on having an initial set of code-switched sentences, which may not be available for all pairs of languages, especially in a low-resource scenario. We believe that it would be interesting to explore the possibility of a cross-lingual approach using our methodology, with English and/or Spanish as pivot languages, that could be useful for transfer knowledge into other less-resourced language pairs.

Finally, as we have pointed out, we are aware of the problems of the automatic metrics that we have used to evaluate the outputs of our models, which do not capture the nuances of our task. In the future, we would like to investigate how to improve this evaluation by designing new methods to automatically evaluate CS generation, focusing on a more linguistic approach able to capture the linguistic and social intricacies of CS.

\section*{Acknowledgments}

This work is supported by the projects DeepMinor (CNS2023-144375) funded by MTDFP/ and by European Union Next GenerationEU/ PRTR; HumanAIze (AIA2025-163322-C61) funded by MICIU/AEI/10.13039/501100011033; and DeepThought (PID2024-159202OB-C21) funded by MICIU/AEI /10.13039/501100011033 and by ERDF, EU. 
Maite Heredia holds a PhD grant from the University of the Basque Country UPV/EHU (PIF23/218).
\nocite{*}

\section*{Bibliographical References}\label{sec:reference}

\bibliographystyle{lrec2026-natbib}
\bibliography{anthology}

\appendix
\section{Prompt-tuning for CS-EN translation}
\label{sec:app-prompt-tuning}
For CS$\rightarrow$EN translation of the LINCE benchmark, we test the prompts in Table \ref{tab:prompts}, combined with 0-, 1- and 5-shot strategies. The prompts include the instructions explained in different ways, including more or less information. 

\begin{table*}[t]
    \renewcommand{\arraystretch}{1.2} 
    \centering
    \begin{tabular}{p{0.95\textwidth}} 
        \hline
        Convert this code-switched phrase to English.\\ \hline 
        Convert this code-switched phrase to English without correcting the original spelling, focus on translating the parts in Spanish.\\ \hline
        \textbf{Convert this code-switched phrase to English. Leave the parts in Spanish as they are, focus on translating the parts in Spanish.}\\\hline 
        Convert this code-switched phrase to English. Directly output the translation and don't correct the original spelling, focus on translating the parts in Spanish.\\ \hline
    \end{tabular}
    \caption{Different prompts that have been used to convert the code-switched instances into English, with different levels of specificity. Final prompt in bold.}
    \label{tab:prompts}
\end{table*}

For the few-shot strategies, the prompt includes the following template at the beginning, alongside a set of manually selected examples that are representative of some phenomena we want to cover in our prompt:

\textit{Here are \{n\} examples of a code-switched text that has been converted to \{lang\}:}

Testing the different prompts, we are able to choose the one whose outputs are closest to our needs, taking into consideration the trade-off between including too little and too much level of specificity in the instructions to the models. 

Regarding the few-shot strategies, we find out that giving some examples to the models results in outputs that are more aligned with the expected output, which is logical, since this allows the models to more faithfully replicate the examples provided. The more examples given, the more the model is able to comply to leaving the punctuation marks as they are and not standardizing the spelling, but also it tends to add more colloquial terms and alternate spellings. Here are the final 5 examples that were selected as they cover the majority of the phenomena that models were observed to struggle with:
\small
\begin{description}[leftmargin=0pt,labelindent=0pt]
\item[\textbf{Input}] cuando me gusta algo nunca lo hay mi fucking size o no tengo el dinero.
\item[\textbf{Output}] when I like something there's never my size or I don't have the money.
\medskip
\item[\textbf{Input}] excelente compartir contigo gracias por tu amistad <user> u rock
\item[\textbf{Output}] excellent sharing with you thank you for your friendship <user> u rock
\medskip
\item[\textbf{Input}] fuhk it tacos de frijol
\item[\textbf{Output}] fuhk it bean tacos
\medskip
\item[\textbf{Input}] <user> como se llama esa app i wanna play it lmfao
\item[\textbf{Output}] <user> what's that app called i wanna play it lmfao
\medskip
\item[\textbf{Input}] i tried putting fake eyelashes on rn lmao me ebarre de glue todo el pinche ojo jajajaja \#osoalmil jajaja
\item[\textbf{Output}] i tried putting fake eyelashes on rn lmao i put glue all over my damn eye hahahaha \#superclumsy hahaha
\end{description}

\section{Post-edition Guidelines}
\label{sec:appendix-post-edition-guidelines}
The original sentence should contain \underline{\textbf{CS}} and be \underline{\textbf{translatable}}. The main reasons to \textbf{remove} an instance altogether are:
\begin{itemize}
    \item If the sentence is very clearly monolingual and the CS has been detected incorrectly (eg, the case of interlingual homographs such as \textit{has}).
    \item When the sentence is bilingual for metalinguistic reasons, because it makes the translation tricky and hard to understand, and in most cases it’s not even CS.
    \item The part that is in the other language is a named entity, such as a title, a name, …
    \item If the code-switched part is not translatable or very hard to translate, probably because it’s a borrowing. Ambiguous and a little bit up to the annotator.
    \item If the tweet is saying the same thing in both languages (making it monolingual doesn’t make sense).
    \item Some instances are tweets that are part of a conversation or thread and taken out of context are very hard to understand/intelligible.
    \item Some tweets are not translatable because of wordplay that doesn’t transfer to monolingual speech.
\end{itemize}
The result should be a \underline{\textbf{monolingual}} sentence that has roughly the \textbf{\underline{same meaning}} as the original sentence. The main reasons to edit a translation are:
\begin{itemize}
    \item If the meaning changes or the model has hallucinated extra information that wasn’t present in the original sentence.
    \item If there are still some words in the Spanish.
    \item Attempts to translate named entities.
    \item Remove “meta comments” from the model about the task.
\end{itemize}

It is not necessary to correct things like:
\begin{itemize}
    \item Punctuation marks.
    \item Different spellings of the same word.
    \item Words of phrases that the model has changed for synonyms.
\end{itemize}

\section{Inter-annotator agreement}
\label{sec:IAA}
\begin{figure*}
    \centering
    \includegraphics[width=0.49\linewidth]{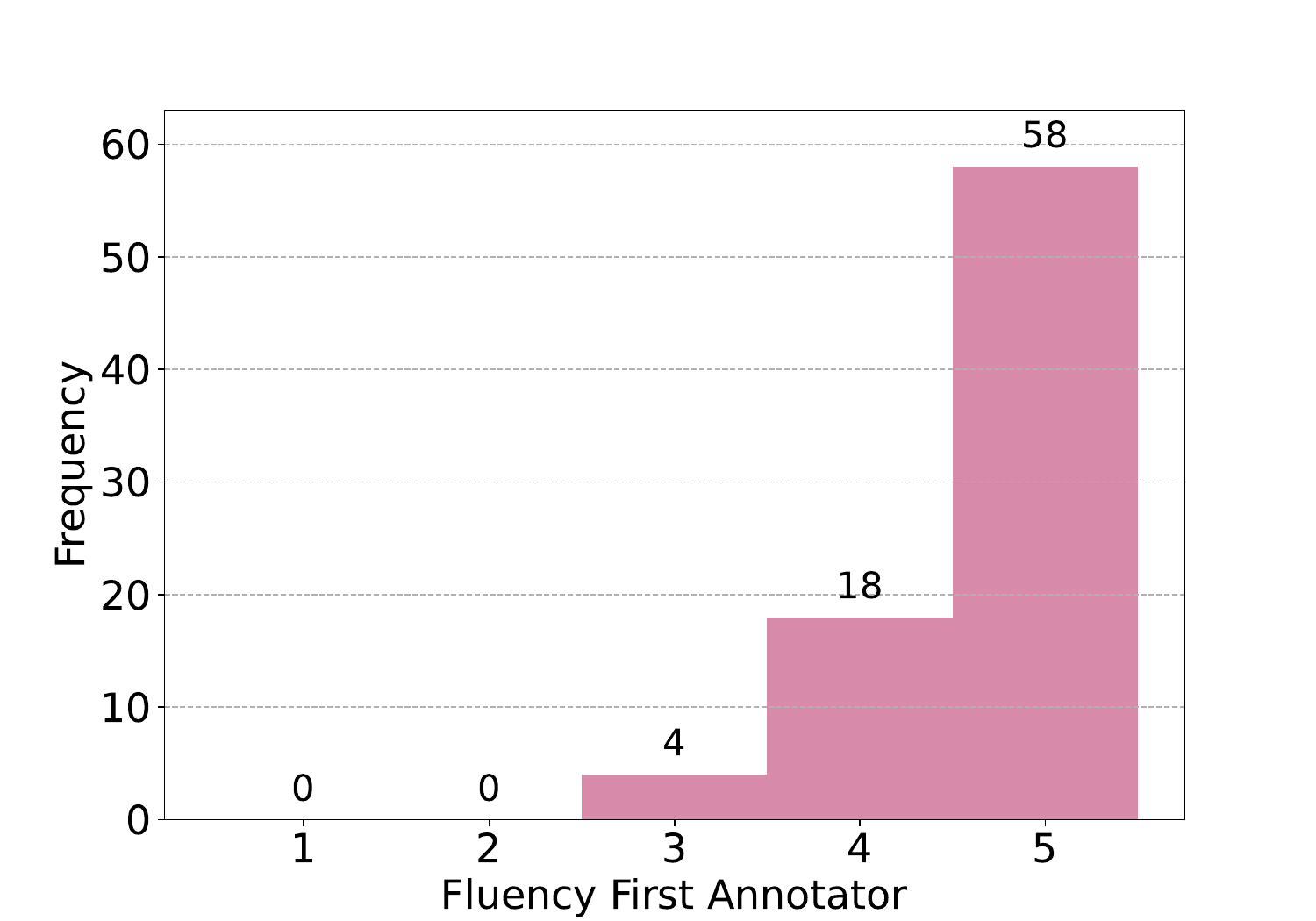}
    \includegraphics[width=0.49\linewidth]{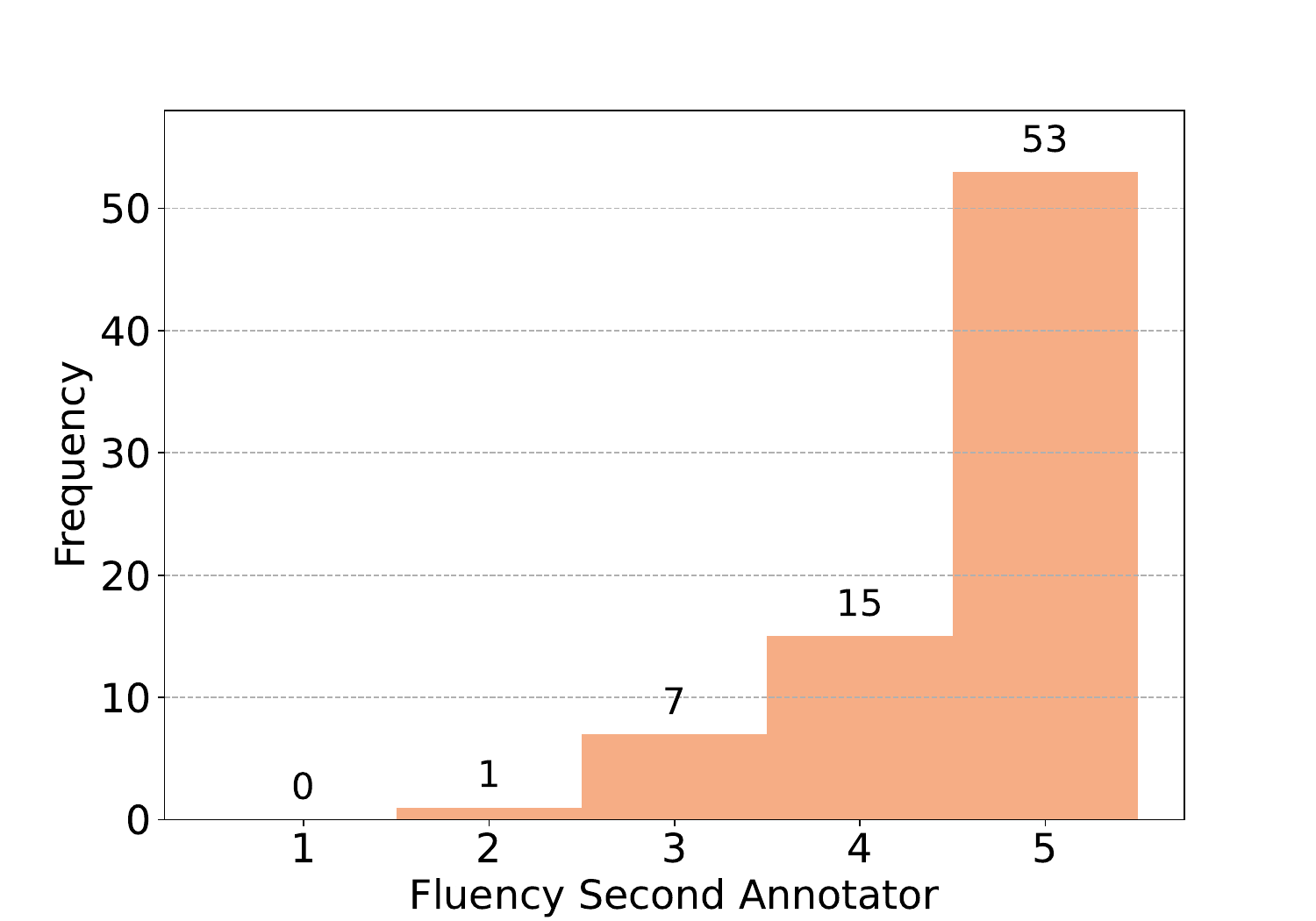}
    \includegraphics[width=0.49\linewidth]{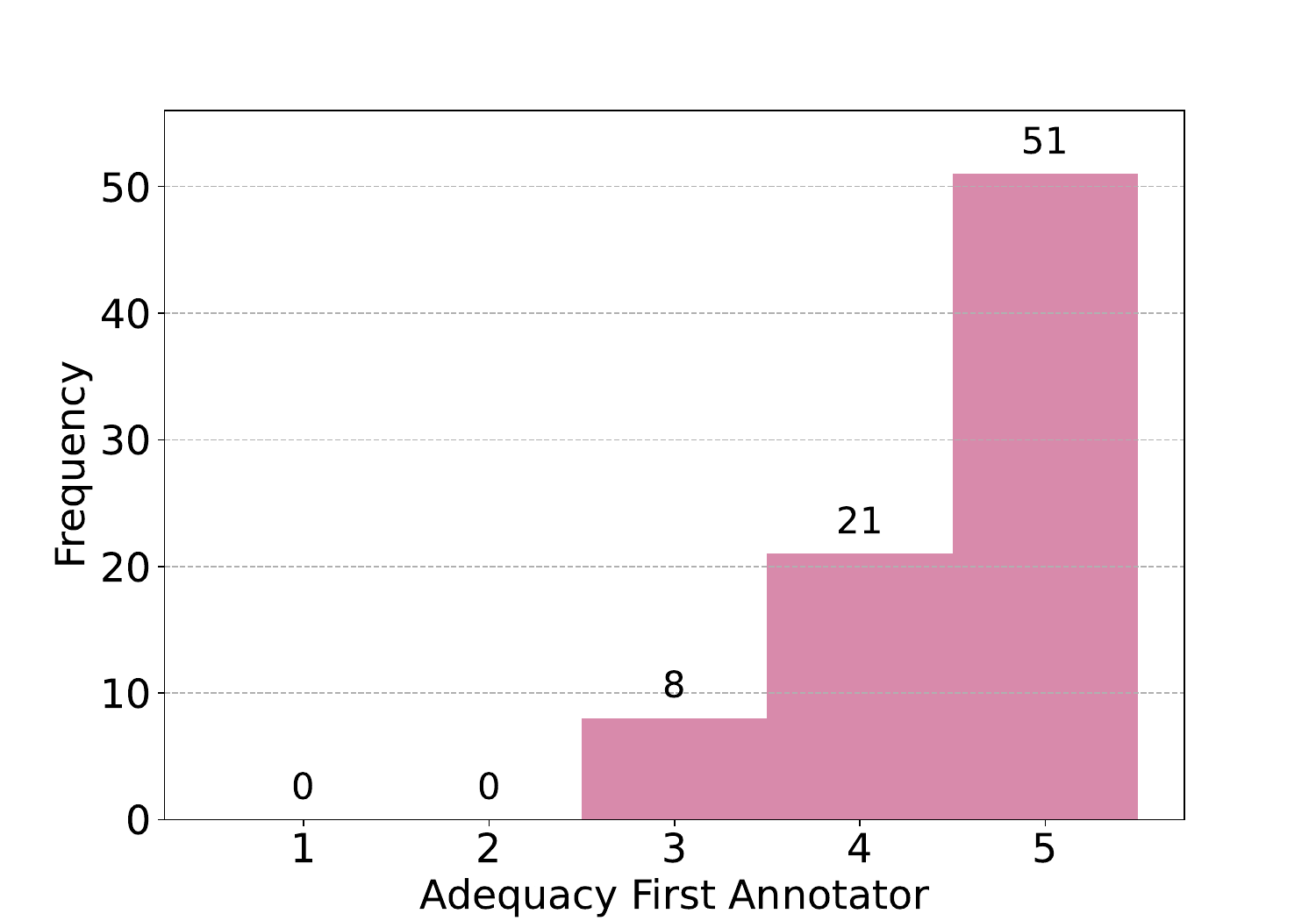}
    \includegraphics[width=0.49\linewidth]{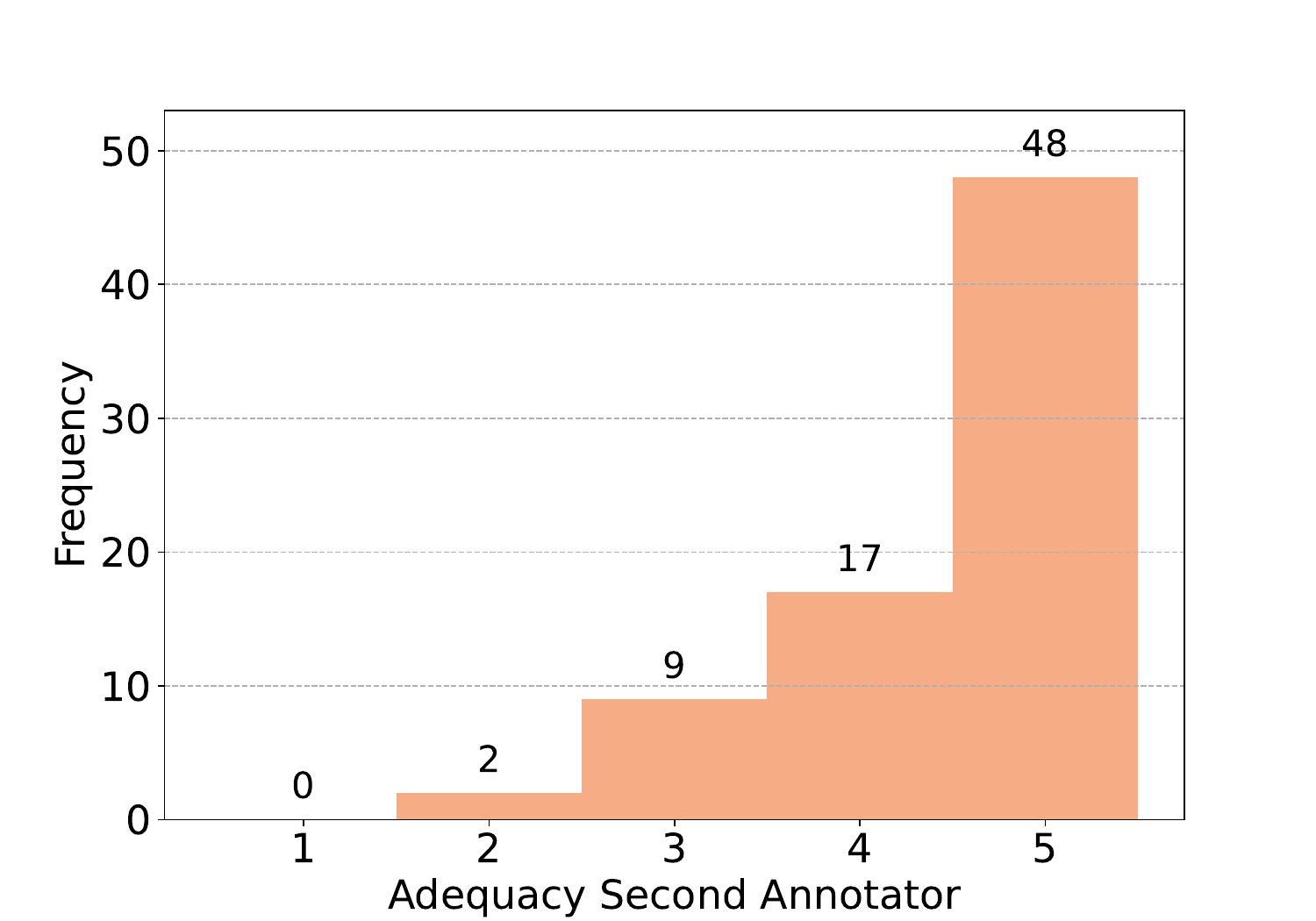}
    \caption{Distribution of Adequacy and Fluency scores per annotator.}
    \label{fig:iaa-distribution}
\end{figure*}
Figure \ref{fig:iaa-distribution} shows the distribution of the scores given by to annotators on the fluency and adequacy of the instances translated by Command R. Although the inter-annotator agreement shows moderate agreement ($\kappa=0.57$), the distributions between the annotators are very similar to each other.

\section{Fine-tuning / Few-shot prompting}
\label{sec:app-ft-prompting}
\begin{table}[t]
  \centering
  \begin{tabular}{p{.45\textwidth}}
    \hline
    \multicolumn{1}{c}{\textbf{Base model}}                                                                                                                                                                                                           \\ \hline
    \small \textit{I want to not work and make money.} = \textit{quiero no trabajar and make money}                                                                                                                                                   \\ \hline
    \multicolumn{1}{c}{\textbf{Instruction-tuned model}}                                                                                                                                                                                              \\
    \cmidrule(lr){1-1}\small \textbf{system prompt}: "\textit{You are a bilingual speaker of English and Spanish. Translate the following English sentence into code-switched text between both languages:}"                                          \\
    \small  \textbf{user}: "\textit{I want to not work and make money.}"                                                                                                                                                                              \\
    \small \textbf{assistant}: "\textit{quiero no trabajar and make money}"                                                                                                                                                                           \\\hline
    \multicolumn{1}{c}{\textbf{Few-shot prompting}}                                                                                                                                                                                                   \\
    \cmidrule(lr){1-1}\small \textbf{system prompt}: "\textit{You are a bilingual speaker of English and Spanish. Translate the following English sentence into code-switched text between both languages. Do not add any comments or explanations:}" \\
    \small  \textbf{user}: Source example $n$                                                                                                                                                                                                         \\
    \small  \textbf{assistant}: Target example $n$                                                                                                                                                                                                    \\
    \small  \textbf{user}: "\textit{I want to not work and make money.}"                                                                                                                                                                              \\
    \small \textbf{assistant}: ...                                                                                                                                                                                                                    \\\hline
      \end{tabular}
      \caption{Examples and format of prompts used for finetuning base and instruction-tuned models and for few-shot prompting} 
  \label{tab:exampl-finet-base}
\end{table}

In Table \ref{tab:exampl-finet-base}, we can see the prompt used for a)fine-tuning of base models; b) fine-tuning of instruction-tuned models; and c) 5-shot prompting. For both fine-tuning settings, at inference time the second part of the prompt that contains the target CS sentence is left blank for the model to complete. For the few-shot prompting approach, the examples are selected randomly among the rest of the instances of the test set.

\section{Pairwise Annotation Guidelines}
\label{sec:annotation-guidelines}
The main objective of this task is two evaluate a pair of sentences that \textbf{should contain code-switching between English and Spanish}.
It should be noted that models have been trained with texts extracted from social media and informal conversations, so \textbf{the outputs of the models are expected to present traits of informality}, such as common typos, that at first should not be considered errors, because they are within the expected behaviour of the models. 
The criteria to choose between both sentences is to be applied \textbf{in the following order}:

\begin{enumerate}
    \item \textbf{Code-switching}
    \begin{enumerate}[label*=\arabic*.]
        \item \textbf{Presence of code-switching}: For a sentence to be a suitable candidate it must have tokens in both languages. A completely monolingual sentence will always be wrong.
        \item \textbf{Naturalness of the code-switching}: A switch between both languages can be unnatural. There are different linguistic constraints. For example, a switch is only possible at a point in a sentence where it does not violate the syntactic rules of either language.
    \end{enumerate}
    \item \textbf{Content and fluency}
    \begin{enumerate}[label*=\arabic*.]
        \item \textbf{Content}: Sentences must have meaning as a whole, they have to be understandable, without extra content disconnected from the rest of the message or abrupt interruptions.
        \item \textbf{Agreement}: Sentences must have the right gender and number agreement.
        \item \textbf{Conjugation}: Verbs have to be correctly conjugated.
    \end{enumerate}
    \item \textbf{Form}: Additional errors that can be used in case none of the above are applicable.
    \begin{enumerate}[label*=\arabic*.]
        \item \textbf{Repetitions} of the same word or phrase.
        \item \textbf{Misspelled words / uncommon typos}
        \item \textbf{Wrong punctuation marks}
        \item \textbf{Extra characters}
    \end{enumerate}
\end{enumerate}

\textbf{Ties are only contemplated in two situations}:
\begin{itemize}
    \item Two sentences that are \textbf{equally wrong}, that is to say, they are both either completely monolingual or unintelligible.
    \item Two sentences that are \textbf{exactly the same} and thus no criteria can be used to break the tie.
\end{itemize}

In case no criteria is applicable to a pair, we ask the annotators to choose their preferred sentence, using their own judgment o additional criteria they might observe in the specific pair of sentences.
\newpage
\section{Error Typology}
\label{sec:appendix-error-typology}
\begin{enumerate}
    \item \textbf{CS errors}
    \begin{enumerate}[label*=\arabic*.]
        \item \textbf{No CS} - the sentence is entirely monolingual.
        \item \textbf{Unnatural CS} - the sentence contains unnatural CS, either due to unnatural switching points, or unnatural register.
        \item \textbf{Repetition in both languages} - the sentence contains the same information repeated in both languages, rather than CS.
    \end{enumerate}
    \item \textbf{Translation errors}
    \begin{enumerate}[label*=\arabic*.]
        \item \textbf{Made-up words} - the words in the output look like English or Spanish but do no actually exist.
        \item \textbf{Wrong translation } - the translation of a word or phrase is incorrect.
        \item \textbf{Wrong conjugation} - a verb is translated with the right lexeme but a seemingly made-up conjugation.
        \item \textbf{Wrong agreement} - there is a mistake in agreement in gender or number.
        \item \textbf{Wrong meaning} - a word or phrase has been translated into a sense that does not fit into the context.
        \item \textbf{Wrong order} - the words are right but they are written in the wrong order.
        \item \textbf{Wrong tense} - the verbal tense is not consistent through the sentence.
        \item \textbf{Unintelligible} - it is not possible to understand the sentence in English nor in Spanish.
        \item \textbf{Instruction misunderstanding} - the task has been misunderstood, e.g., the model makes a "comment" about the content of the output or explains a word. 
    \end{enumerate}
    \item \textbf{Format errors}
    \begin{enumerate}[label*=\arabic*.]
        \item \textbf{Extra words }- the sentence contains seemingly random extra words that do not affect its meaning.
        \item \textbf{Extra characters} - the sentence contains more non-word characters than the original, e.g., `???' instead of `??'.
        \item \textbf{Hallucinations} - the sentence contains new words or phrases not derived from the original text.
        \item \textbf{Start over} - the sentence is finalized, but the model begins a second translation of the same sentence.
        \item \textbf{Duplications} - some words or phrases of the sentence are duplicated.

    \end{enumerate}
\end{enumerate}
\begin{table}[t!]
    \centering
    \begin{tabular}{p{0.9\columnwidth}} \hline 
        \textbf{developer}: "You are a helpful bilingual system that knows how to code-switch between English and Spanish and how to distinguish natural sentences. Your only job is to judge sentences and output a verdict A, B or T."\\
        \textbf{user}:"Which one of the next two automatically generated sentences with code-switching is more natural? The most important criterion is that the sentences must have code-switching to even be considered eligible. Two sentences can be tied if they are equally wrong.\\
        A: \{s1\}\\
        B: \{s2\}\\
        Answer(A/B/T):"
        \\ \hline
    \end{tabular}\par
        \caption{Prompt used for GPT to act as a judge.}
        \label{tab:gpt-judge-prompts}
\end{table}
\section{Implementation of GPT as judge}
\label{sec:appendix-gpt-judge}

For the implementation of GPT as judge, the developer and user prompts in Table~\ref{tab:gpt-judge-prompts} have been used to prompt GPT-4o.
To calculate the scores, we first check the answers the directly contain the desired format, "A", "B or "T", which are the most common. For the rest of the outputs that did not follow this format, it is possible to extract the labels using simple regular expressions. Then, the scores are calculated just like the human score: every time a model is voted, it gets $1$ point, and the loser gets $0$ points; in case of ties, both models get $0.5$ points each.
\end{document}